\documentclass{article}

\PassOptionsToPackage{numbers, compress}{natbib}

\usepackage[sglblindworkshop, final]{latinx_2025}

\usepackage{amsmath}
\usepackage{graphicx}

\usepackage[utf8]{inputenc} % allow utf-8 input
\usepackage[T1]{fontenc}    % use 8-bit T1 fonts
\usepackage{hyperref}       % hyperlinks
\usepackage{url}            % simple URL typesetting
\usepackage{booktabs}       % professional-quality tables
\usepackage{amsfonts}       % blackboard math symbols
\usepackage{nicefrac}       % compact symbols for 1/2, etc.
\usepackage{microtype}      % microtypography
\usepackage{xcolor}         % colors
\usepackage{float}
\title{Extending NGU to Multi-Agent RL: A Preliminary Study}

\author{
    \textbf{Juan Hernandez}\textsuperscript{1,3}\thanks{Equal contribution.}   , 
    \textbf{Diego Fernández}\textsuperscript{1,3}\footnotemark[1]   , 
    \textbf{Manuel Cifuentes}\textsuperscript{1,3}\footnotemark[1]   ,\\
    \textbf{Denis Parra}\textsuperscript{1,2,3}, 
    \textbf{Rodrigo Toro Icarte}\textsuperscript{1,3}\\[4pt]
    \textsuperscript{1}Department of Computer Science, Pontifical Catholic University of Chile.\\
    \textsuperscript{2}Millennium Institute for Intelligent Healthcare Engineering (iHEALTH), Chile.\\
    \textsuperscript{3}National Center for Artificial Intelligence (CENIA), Chile.\\[4pt]
    \texttt{\{juan\_manuel1402, diegofpdt, mecifuentes, dparras, rntoro\}@uc.cl}
}

\begin{document}

\maketitle

\begin{abstract}
The Never Give Up (NGU) algorithm has proven effective in reinforcement learning tasks with sparse rewards by combining episodic novelty and intrinsic motivation. In this work, we extend NGU to multi-agent environments and evaluate its performance in the simple\_tag environment from the PettingZoo suite. Compared to a multi-agent DQN baseline, NGU achieves moderately higher returns and more stable learning dynamics. Building on this, we investigate three design choices: (1) shared replay buffer versus individual replay buffers, (2) sharing episodic novelty among agents using different $k$ thresholds, and (3) using heterogeneous values of the $\beta$ parameter. Our results show that NGU with a shared replay buffer yields the best performance and stability, highlighting that the gains come from combining NGU’s intrinsic exploration with experience sharing. Sharing novelty produces comparable performance when $k=1$, but degrades learning for larger $k$ values. Finally, heterogeneous $\beta$ values do not improve over a small common value. These findings suggest that NGU can be effectively applied in multi-agent settings when experiences are shared and intrinsic exploration signals are carefully tuned.
\end{abstract}

\section{Introduction}

Reinforcement Learning (RL) has achieved success in diverse domains such as game-playing \citep{Silver_2016}, recommender systems \citep{10.1145/3366423.3380130}, and medicine \citep{zhou2021deepreinforcementlearningmedical}. A key milestone in this history of success was the Atari 2600 benchmark from the Arcade Learning Environment \citep{Bellemare_2013}, which popularized the Deep Q-Network (DQN) after achieving human-level performance on several games \citep{mnih2013playingatarideepreinforcement}. However, DQN still struggled in sparse-reward environments such as Montezuma’s Revenge, where rewards are extremely delayed and exploration is especially difficult, motivating a line of research on advanced exploration methods \citep{salimans2018learningmontezumasrevengesingle, ecoffet2021goexplorenewapproachhardexploration, tang2017explorationstudycountbasedexploration, ijcai2020p290}.

The Never Give Up (NGU) algorithm \citep{badia2020uplearningdirectedexploration} handles sparse rewards through intrinsic motivation, encouraging agents to explore novel states by combining an episodic novelty (computed in an embedding space trained by an inverse dynamics model) with a life-long novelty modulator based on Random Network Distillation (RND) \citep{burda2018explorationrandomnetworkdistillation}. This design allows NGU to balance within-episode exploration and cross-episode discovery, enabling agents to achieve state-of-the-art results on previously unsolved sparse-reward Atari games such as Montezuma’s Revenge and Pitfall.

In the context of Multi-Agent Reinforcement Learning (MARL), sparse reward environments pose even greater challenges due to issues such as credit assignment, non-stationarity, and the need for coordinated exploration \citep{pmlr-v202-liu23ac, ijcai2024p32}.

Recent works in multi-agent reinforcement learning have explored intrinsic-motivation mechanisms to address the challenges of sparse rewards. For instance, \textbf{EMC} (Episodic Multi-agent reinforcement learning with Curiosity-driven exploration) leverages the prediction error of factorized Q-values as a curiosity signal and uses episodic memory to reinforce informative trajectories \citep{zheng2021episodicmultiagentreinforcementlearning}. Similarly, \textbf{MACE} (Multi-Agent Coordinated Exploration) allows decentralized agents to approximate global novelty by sharing local novelty estimates and introduces a hindsight-based intrinsic reward based on weighted mutual information \citep{jiang2024settlingdecentralizedmultiagentcoordinated}.

Despite these advances, as well as others \citep{na2024efficientepisodicmemoryutilization, iqbal2021coordinatedexplorationintrinsicrewards, NEURIPS2019_07a9d3fe}, existing methods often introduce additional architectural complexity, computational overhead, or rely on carefully designed intrinsic signals that may not generalize across tasks. To address this, we revisit the simpler yet powerful NGU framework and investigate how its mechanisms of exploration can be adapted to multi-agent environments.

In extending NGU beyond the single-agent setting, we deliberately exclude components such as Random Network Distillation (RND) and Universal Value Function Approximators (UVFA) \citep{pmlr-v37-schaul15}, which add complexity and computational cost. Instead, we focus on the core components of NGU that drive exploration: the inverse dynamics model for representation learning, the embedding network, the episodic memory, and the computation of intrinsic rewards based on state novelty. This setup preserves the essence of NGU while making it feasible to study its adaptation in MARL enviroments.

\begin{figure}[t]
    \centering
    \fbox{\includegraphics[width=0.25\linewidth]{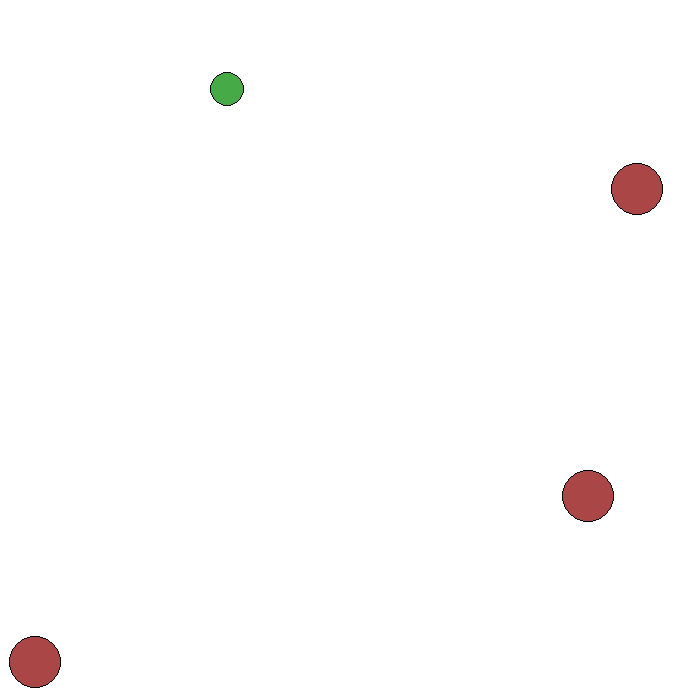}}
    \caption{The simple\_tag environment from the PettingZoo suite. Multiple pursuers (red) cooperate to capture an evader (blue) in a bounded 2D arena.}
    \label{fig:env}
\end{figure}

To provide a baseline, we compare our method against a multi-agent DQN trained under the same conditions and parameters in the simple\_tag environment from the PettingZoo suite \citep{terry2021pettingzoogymmultiagentreinforcement}. This pursuit–evasion task involves multiple pursuers cooperating to capture an evader, as illustrated in Figure~\ref{fig:env}. While the baseline is able to solve the task occasionally, our Multi-NGU approach achieves moderately higher average returns and exhibits more stable learning dynamics across training runs. The implementation of our method is publicly available at \citep{hernandez2025extending}.

Building on these findings, we investigate three design choices for extending NGU to the multi-agent setting: (1) whether to pool trajectories in a shared replay buffer or to maintain individual buffers per agent, (2) whether to share episodic novelty across agents by varying the $k$ threshold that determines when a state is no longer considered novel, and (3) whether to assign heterogeneous values of the intrinsic-extrinsic trade-off parameter $\beta$ across agents instead of using a common small value. 

\paragraph{Contributions.} Our main contributions are a direct extension of NGU to cooperative multi-agent environments with sparse rewards; an analysis of three design choices: replay sharing, novelty sharing, and heterogeneous $\beta$; empirical evidence that Multi-NGU improves stability and returns over Multi-DQN; and clear directions for future research on extending NGU to MARL.

\section{Single agent NGU}

The NGU algorithm combines multiple sources of intrinsic motivation to encourage exploration in sparse-reward environments. In our work, we adopt only its essential components while omitting Random Network Distillation (RND) and Universal Value Function Approximators (UVFA) for simplicity and computational resources.

Specifically, we retain an embedding network, which encodes raw observations into a compact representation space, and an inverse dynamics model, which is trained to predict the action between consecutive states in this space. Together, these modules support an episodic memory that stores embeddings within each episode and allows novelty to be measured through $k$-nearest neighbors. Based on this measure, an intrinsic reward is computed so that states considered novel with respect to the episodic memory yield higher motivation. 

This reduced but faithful formulation preserves NGU’s core mechanism of rewarding novelty while keeping the method computationally tractable for multi-agent experiments.

\section{Extending NGU to MARL}

First, we experiment with a Multi-NGU approach. Each agent $i \in \{1,\dots,N\}$ has its own Q-network, its own embedding network, episodic memory, and intrinsic reward. Thus, exploration signals remain individualized, even though agents interact in a shared environment.

Let $\phi: \mathcal{S} \rightarrow \mathbb{R}^d$ denote the embedding network, trained via an inverse dynamics loss to predict the action $a_t$ given consecutive embeddings $(\phi(s_t), \phi(s_{t+1}))$. At each timestep, agent $i$ computes the episodic novelty of its next state embedding $\phi(s_{t+1}^i)$ with respect to its episodic memory $\mathcal{M}_i$:

\begin{equation}
r^{\text{intrinsic}}_{t,i} = f\!\left(\phi(s_{t+1}^i), \; \mathcal{M}_i \right),
\end{equation}

where $f$ is a $k$-nearest neighbor distance function, and $\mathcal{M}_i$ is the episodic memory buffer storing embeddings observed by agent $i$ within the current episode.

The total reward used for learning is the combination of the extrinsic reward $r^{\text{extrinsic}}_t$ and the intrinsic novelty reward scaled by parameter $\beta_i$:

\begin{equation}
r_{t,i} = r^{\text{extrinsic}}_t + \beta_i \, r^{\text{intrinsic}}_{t,i}.
\end{equation}

Building on this base formulation, we investigate three design choices to adapt NGU more effectively to the multi-agent setting.

\paragraph{Shared replay buffer.} 
Instead of each agent maintaining an independent buffer, all experiences are pooled into a centralized replay. This improves sample efficiency and reduces non-stationarity, as agents benefit from the trajectories of others.

\paragraph{Shared novelty.} 
We also test sharing novelty across agents. A state embedding becomes ``non-novel'' for everyone once visited by $k$ different agents. To detect similarity, we use cosine similarity between projected embeddings, so that states with high cosine overlap are treated as already known.

\paragraph{Heterogeneous $\beta$.} 
Finally, we vary the intrinsic/extrinsic trade-off parameter $\beta$ across agents (e.g., $\{0.1, 0.2, 0.4\}$) instead of fixing a small common value. The idea is to diversify roles, letting some agents emphasize exploration and others exploitation.

\section{Experimental Setup}
We evaluate our approach in the simple\_tag\_v3 environment from the PettingZoo suite, a standard multi-agent RL library that facilitates reproducibility. Rewards are sparse and shared: when any pursuer tags the evader, all pursuers receive the same reward.

Each pursuer is implemented as a DQN agent augmented with NGU components. The evader follows the default heuristic policy provided by PettingZoo.

We experiment with two scenarios: one without a shared replay buffer and one with a shared replay buffer. In each scenario, we evaluate four configurations under identical conditions: the Multi-DQN baseline, Multi-DQN augmented with NGU, the novelty sharing variant, and the heterogeneous $\beta$ variant. This design enables us to assess the contribution of NGU and its design choices consistently across both replay buffer settings.

All agents have the same network architecture and hyperparameters across conditions, ensuring comparability. The full hyperparameter configuration is reported in the Appendix~\ref{sec:hyperparams}. Unless otherwise stated, we use $\beta = 0.1$; the heterogeneous $\beta$ variant is the only exception. We run every configuration with identical random seeds for fairness. Each experiment consists of $200{,}000$ timesteps, and we perform $15$ independent runs per configuration.

\section{Results}

\begin{figure}[htpb]
    \centering
    \includegraphics[width=0.95\linewidth]{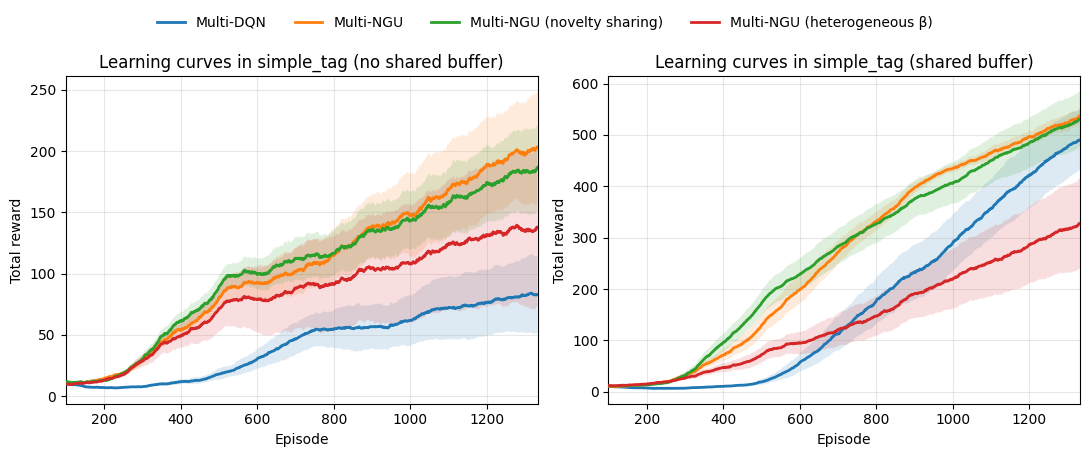}
    \caption{Learning curves of pursuers in the \texttt{simple\_tag} environment. Results are averaged over 15 runs with smoothed returns (window=100), and the shaded regions indicate the 95\% confidence interval. The left panel corresponds to training without a shared replay buffer, while the right panel shows results with buffer sharing.}
    \label{fig:results}
\end{figure}

Preliminary, the heterogeneous $\beta$ variant was evaluated with values $\{0.1, 0.2, 0.4\}$, which are the ones reported in the figures above. Larger or smaller values led to poor performance (see Appendix~\ref{sec:appendix_hetero_beta}). Similarly, in the multi-novelty setting we tested different thresholds $k$, finding that $k=1$ yields the best results. The curves shown in Figure~\ref{fig:results} correspond to this case, while experiments with $k>1$ resulted in degraded performance (see Appendix~\ref{sec:appendix_multinovelty}).

When agents rely on individual replay buffers, the Multi-DQN baseline shows slow and unstable learning, achieving only modest returns. In contrast, incorporating NGU consistently boosts performance, with higher average returns and smoother learning dynamics. Variants such as novelty sharing and heterogeneous $\beta$ do not surpass standard NGU, but still outperform the baseline, highlighting that intrinsic motivation provided by NGU is beneficial even when agents learn from isolated experiences.

Under the shared buffer setting, the advantages of NGU become even clearer. Multi-DQN benefits from the additional data but still lags behind the NGU variants, which achieve substantially higher returns and more stable curves. Standard NGU delivers the strongest results, while novelty sharing performs comparably and heterogeneous $\beta$ again trails behind. Interestingly, multi-novelty with $k=1$ and multi-NGU with a shared buffer show some conceptual similarity, since sharing novelty can be seen as analogous to sharing experience: in both cases, the computation of intrinsic rewards is effectively reduced when an experience has already been observed. However, their learning dynamics differ from what might be expected: multi-novelty exhibits a faster increase at the beginning of training, but at some point standard multi-NGU surpasses it and achieves higher long-term returns. Overall, these results reinforce that NGU’s intrinsic motivation mechanisms significantly improve multi-agent learning outcomes across conditions.

\section{Future Work}

Since this work focuses on a single environment and a single algorithm (DQN), there remain many opportunities for future research. First, it would be valuable to evaluate different configurations within the simple\_tag environment. Beyond this domain, the approach could be tested in a broader range of multi-agent settings, from simpler coordination tasks to large-scale competitive scenarios, in order to better assess its generality and scalability. Future research could also explore alternative algorithms such as value decomposition methods (e.g., VDN \citep{sunehag2017valuedecompositionnetworkscooperativemultiagent}, QMIX \citep{rashid2018qmixmonotonicvaluefunction}) or policy-gradient approaches (e.g., MADDPG \citep{lowe2020multiagentactorcriticmixedcooperativecompetitive}, MAPPO \cite{yu2022surprisingeffectivenessppocooperative}). Finally, to strengthen the empirical analysis, it will be important to compare Multi-NGU against stronger benchmarks, including large-scale environments such as the StarCraft Multi-Agent Challenge \citep{samvelyan2019starcraftmultiagentchallenge}, providing a more comprehensive evaluation of its robustness and scalability.

\begin{ack}
This work was partially supported by the National Center for Artificial Intelligence CENIA FB210017, Basal ANID
\end{ack}

\bibliographystyle{plainnat}
\bibliography{references}

@article{Silver_2016,
  author = {Silver, David and Huang, Aja and Maddison, Chris J. and Guez, Arthur and Sifre, Laurent and van den Driessche, George and Schrittwieser, Julian and Antonoglou, Ioannis and Panneershelvam, Veda and Lanctot, Marc and Dieleman, Sander and Grewe, Dominik and Nham, John and Kalchbrenner, Nal and Sutskever, Ilya and Lillicrap, Timothy and Leach, Madeleine and Kavukcuoglu, Koray and Graepel, Thore and Hassabis, Demis},
  doi = {10.1038/nature16961},
  journal = {Nature},
  month = jan,
  number = 7587,
  pages = {484--489},
  publisher = {Nature Publishing Group},
  title = {Mastering the Game of {Go} with Deep Neural Networks and Tree Search},
  volume = 529,
  year = 2016
}

@inproceedings{10.1145/3366423.3380130,
author = {Ma, Jiaqi and Zhao, Zhe and Yi, Xinyang and Yang, Ji and Chen, Minmin and Tang, Jiaxi and Hong, Lichan and Chi, Ed H.},
title = {Off-policy Learning in Two-stage Recommender Systems},
year = {2020},
isbn = {9781450370233},
publisher = {Association for Computing Machinery},
address = {New York, NY, USA},
url = {https://doi.org/10.1145/3366423.3380130},
doi = {10.1145/3366423.3380130},
booktitle = {Proceedings of The Web Conference 2020},
pages = {463–473},
numpages = {11},
location = {Taipei, Taiwan},
series = {WWW '20}
}

@misc{zhou2021deepreinforcementlearningmedical,
      title={Deep reinforcement learning in medical imaging: A literature review}, 
      author={S. Kevin Zhou and Hoang Ngan Le and Khoa Luu and Hien V. Nguyen and Nicholas Ayache},
      year={2021},
      eprint={2103.05115},
      archivePrefix={arXiv},
      primaryClass={eess.IV},
      url={https://arxiv.org/abs/2103.05115}, 
}

@article{Bellemare_2013,
   title={The Arcade Learning Environment: An Evaluation Platform for General Agents},
   volume={47},
   ISSN={1076-9757},
   url={http://dx.doi.org/10.1613/jair.3912},
   DOI={10.1613/jair.3912},
   journal={Journal of Artificial Intelligence Research},
   publisher={AI Access Foundation},
   author={Bellemare, M. G. and Naddaf, Y. and Veness, J. and Bowling, M.},
   year={2013},
   month=jun, pages={253–279} }

@misc{mnih2013playingatarideepreinforcement,
      title={Playing Atari with Deep Reinforcement Learning}, 
      author={Volodymyr Mnih and Koray Kavukcuoglu and David Silver and Alex Graves and Ioannis Antonoglou and Daan Wierstra and Martin Riedmiller},
      year={2013},
      eprint={1312.5602},
      archivePrefix={arXiv},
      primaryClass={cs.LG},
      url={https://arxiv.org/abs/1312.5602}, 
}

@misc{salimans2018learningmontezumasrevengesingle,
      title={Learning Montezuma's Revenge from a Single Demonstration}, 
      author={Tim Salimans and Richard Chen},
      year={2018},
      eprint={1812.03381},
      archivePrefix={arXiv},
      primaryClass={cs.LG},
      url={https://arxiv.org/abs/1812.03381}, 
}

@misc{ecoffet2021goexplorenewapproachhardexploration,
      title={Go-Explore: a New Approach for Hard-Exploration Problems}, 
      author={Adrien Ecoffet and Joost Huizinga and Joel Lehman and Kenneth O. Stanley and Jeff Clune},
      year={2021},
      eprint={1901.10995},
      archivePrefix={arXiv},
      primaryClass={cs.LG},
      url={https://arxiv.org/abs/1901.10995}, 
}

@misc{tang2017explorationstudycountbasedexploration,
      title={\#Exploration: A Study of Count-Based Exploration for Deep Reinforcement Learning}, 
      author={Haoran Tang and Rein Houthooft and Davis Foote and Adam Stooke and Xi Chen and Yan Duan and John Schulman and Filip De Turck and Pieter Abbeel},
      year={2017},
      eprint={1611.04717},
      archivePrefix={arXiv},
      primaryClass={cs.AI},
      url={https://arxiv.org/abs/1611.04717}, 
}

@inproceedings{ijcai2020p290,
  title     = {Potential Driven Reinforcement Learning for Hard Exploration Tasks},
  author    = {Zhao, Enmin and Deng, Shihong and Zang, Yifan and Kang, Yongxin and Li, Kai and Xing, Junliang},
  booktitle = {Proceedings of the Twenty-Ninth International Joint Conference on
               Artificial Intelligence, {IJCAI-20}},
  publisher = {International Joint Conferences on Artificial Intelligence Organization},
  editor    = {Christian Bessiere},
  pages     = {2096--2102},
  year      = {2020},
  month     = {7},
  note      = {Main track},
  doi       = {10.24963/ijcai.2020/290},
  url       = {https://doi.org/10.24963/ijcai.2020/290},
}

@misc{badia2020uplearningdirectedexploration,
      title={Never Give Up: Learning Directed Exploration Strategies}, 
      author={Adrià Puigdomènech Badia and Pablo Sprechmann and Alex Vitvitskyi and Daniel Guo and Bilal Piot and Steven Kapturowski and Olivier Tieleman and Martín Arjovsky and Alexander Pritzel and Andew Bolt and Charles Blundell},
      year={2020},
      eprint={2002.06038},
      archivePrefix={arXiv},
      primaryClass={cs.LG},
      url={https://arxiv.org/abs/2002.06038}, 
}

@misc{burda2018explorationrandomnetworkdistillation,
      title={Exploration by Random Network Distillation}, 
      author={Yuri Burda and Harrison Edwards and Amos Storkey and Oleg Klimov},
      year={2018},
      eprint={1810.12894},
      archivePrefix={arXiv},
      primaryClass={cs.LG},
      url={https://arxiv.org/abs/1810.12894}, 
}

@InProceedings{pmlr-v202-liu23ac,
  title = 	 {Lazy Agents: A New Perspective on Solving Sparse Reward Problem in Multi-agent Reinforcement Learning},
  author =       {Liu, Boyin and Pu, Zhiqiang and Pan, Yi and Yi, Jianqiang and Liang, Yanyan and Zhang, D.},
  booktitle = 	 {Proceedings of the 40th International Conference on Machine Learning},
  pages = 	 {21937--21950},
  year = 	 {2023},
  editor = 	 {Krause, Andreas and Brunskill, Emma and Cho, Kyunghyun and Engelhardt, Barbara and Sabato, Sivan and Scarlett, Jonathan},
  volume = 	 {202},
  series = 	 {Proceedings of Machine Learning Research},
  month = 	 {23--29 Jul},
  publisher =    {PMLR},
  url = 	 {https://proceedings.mlr.press/v202/liu23ac.html}
}

@inproceedings{ijcai2024p32,
  title     = {Population-Based Diverse Exploration for Sparse-Reward Multi-Agent Tasks},
  author    = {Xu, Pei and Zhang, Junge and Huang, Kaiqi},
  booktitle = {Proceedings of the Thirty-Third International Joint Conference on
               Artificial Intelligence, {IJCAI-24}},
  publisher = {International Joint Conferences on Artificial Intelligence Organization},
  editor    = {Kate Larson},
  pages     = {283--291},
  year      = {2024},
  month     = {8},
  note      = {Main Track},
  doi       = {10.24963/ijcai.2024/32},
  url       = {https://doi.org/10.24963/ijcai.2024/32},
}

@misc{zheng2021episodicmultiagentreinforcementlearning,
      title={Episodic Multi-agent Reinforcement Learning with Curiosity-Driven Exploration}, 
      author={Lulu Zheng and Jiarui Chen and Jianhao Wang and Jiamin He and Yujing Hu and Yingfeng Chen and Changjie Fan and Yang Gao and Chongjie Zhang},
      year={2021},
      eprint={2111.11032},
      archivePrefix={arXiv},
      primaryClass={cs.LG},
      url={https://arxiv.org/abs/2111.11032}, 
}

@misc{jiang2024settlingdecentralizedmultiagentcoordinated,
      title={Settling Decentralized Multi-Agent Coordinated Exploration by Novelty Sharing}, 
      author={Haobin Jiang and Ziluo Ding and Zongqing Lu},
      year={2024},
      eprint={2402.02097},
      archivePrefix={arXiv},
      primaryClass={cs.MA},
      url={https://arxiv.org/abs/2402.02097}, 
}

@misc{na2024efficientepisodicmemoryutilization,
      title={Efficient Episodic Memory Utilization of Cooperative Multi-Agent Reinforcement Learning}, 
      author={Hyungho Na and Yunkyeong Seo and Il-chul Moon},
      year={2024},
      eprint={2403.01112},
      archivePrefix={arXiv},
      primaryClass={cs.LG},
      url={https://arxiv.org/abs/2403.01112}, 
}

@InProceedings{pmlr-v37-schaul15,
  title = 	 {Universal Value Function Approximators},
  author = 	 {Schaul, Tom and Horgan, Daniel and Gregor, Karol and Silver, David},
  booktitle = 	 {Proceedings of the 32nd International Conference on Machine Learning},
  pages = 	 {1312--1320},
  year = 	 {2015},
  editor = 	 {Bach, Francis and Blei, David},
  volume = 	 {37},
  series = 	 {Proceedings of Machine Learning Research},
  address = 	 {Lille, France},
  month = 	 {07--09 Jul},
  publisher =    {PMLR},
  url = 	 {https://proceedings.mlr.press/v37/schaul15.html}
}

@misc{terry2021pettingzoogymmultiagentreinforcement,
      title={PettingZoo: Gym for Multi-Agent Reinforcement Learning}, 
      author={J. K. Terry and Benjamin Black and Nathaniel Grammel and Mario Jayakumar and Ananth Hari and Ryan Sullivan and Luis Santos and Rodrigo Perez and Caroline Horsch and Clemens Dieffendahl and Niall L. Williams and Yashas Lokesh and Praveen Ravi},
      year={2021},
      eprint={2009.14471},
      archivePrefix={arXiv},
      primaryClass={cs.LG},
      url={https://arxiv.org/abs/2009.14471}, 
}

@misc{iqbal2021coordinatedexplorationintrinsicrewards,
      title={Coordinated Exploration via Intrinsic Rewards for Multi-Agent Reinforcement Learning}, 
      author={Shariq Iqbal and Fei Sha},
      year={2021},
      eprint={1905.12127},
      archivePrefix={arXiv},
      primaryClass={cs.LG},
      url={https://arxiv.org/abs/1905.12127}, 
}

@inproceedings{NEURIPS2019_07a9d3fe,
 author = {Du, Yali and Han, Lei and Fang, Meng and Liu, Ji and Dai, Tianhong and Tao, Dacheng},
 booktitle = {Advances in Neural Information Processing Systems},
 editor = {H. Wallach and H. Larochelle and A. Beygelzimer and F. d\textquotesingle Alch\'{e}-Buc and E. Fox and R. Garnett},
 pages = {},
 publisher = {Curran Associates, Inc.},
 title = {LIIR: Learning Individual Intrinsic Reward in Multi-Agent Reinforcement Learning},
 url = {https://proceedings.neurips.cc/paper_files/paper/2019/file/07a9d3fed4c5ea6b17e80258dee231fa-Paper.pdf},
 volume = {32},
 year = {2019}
}

@misc{samvelyan2019starcraftmultiagentchallenge,
      title={The StarCraft Multi-Agent Challenge}, 
      author={Mikayel Samvelyan and Tabish Rashid and Christian Schroeder de Witt and Gregory Farquhar and Nantas Nardelli and Tim G. J. Rudner and Chia-Man Hung and Philip H. S. Torr and Jakob Foerster and Shimon Whiteson},
      year={2019},
      eprint={1902.04043},
      archivePrefix={arXiv},
      primaryClass={cs.LG},
      url={https://arxiv.org/abs/1902.04043}, 
}

@misc{sunehag2017valuedecompositionnetworkscooperativemultiagent,
      title={Value-Decomposition Networks For Cooperative Multi-Agent Learning}, 
      author={Peter Sunehag and Guy Lever and Audrunas Gruslys and Wojciech Marian Czarnecki and Vinicius Zambaldi and Max Jaderberg and Marc Lanctot and Nicolas Sonnerat and Joel Z. Leibo and Karl Tuyls and Thore Graepel},
      year={2017},
      eprint={1706.05296},
      archivePrefix={arXiv},
      primaryClass={cs.AI},
      url={https://arxiv.org/abs/1706.05296}, 
}

@misc{rashid2018qmixmonotonicvaluefunction,
      title={QMIX: Monotonic Value Function Factorisation for Deep Multi-Agent Reinforcement Learning}, 
      author={Tabish Rashid and Mikayel Samvelyan and Christian Schroeder de Witt and Gregory Farquhar and Jakob Foerster and Shimon Whiteson},
      year={2018},
      eprint={1803.11485},
      archivePrefix={arXiv},
      primaryClass={cs.LG},
      url={https://arxiv.org/abs/1803.11485}, 
}

@misc{lowe2020multiagentactorcriticmixedcooperativecompetitive,
      title={Multi-Agent Actor-Critic for Mixed Cooperative-Competitive Environments}, 
      author={Ryan Lowe and Yi Wu and Aviv Tamar and Jean Harb and Pieter Abbeel and Igor Mordatch},
      year={2020},
      eprint={1706.02275},
      archivePrefix={arXiv},
      primaryClass={cs.LG},
      url={https://arxiv.org/abs/1706.02275}, 
}

@misc{yu2022surprisingeffectivenessppocooperative,
      title={The Surprising Effectiveness of PPO in Cooperative, Multi-Agent Games}, 
      author={Chao Yu and Akash Velu and Eugene Vinitsky and Jiaxuan Gao and Yu Wang and Alexandre Bayen and Yi Wu},
      year={2022},
      eprint={2103.01955},
      archivePrefix={arXiv},
      primaryClass={cs.LG},
      url={https://arxiv.org/abs/2103.01955}, 
}

@misc{hernandez2025extending,
  title        = {Extending NGU to Multi-Agent RL: A Preliminary Study},
  author       = {Juan Hernandez and Diego Fernández and Manuel Cifuentes and Denis Parra and Rodrigo Toro Icarte},
  year         = {2025},
  note         = {GitHub repository},
  howpublished = {\url{https://github.com/JuanHernandez-uc/Extending-NGU-to-MARL}} 
}

\newpage
\appendix

\section{Hyperparameter Selection}
\label{sec:hyperparams}

\begin{table}[H]
\centering
\caption{Base hyperparameters used for Multi-DQN and Multi-NGU. Variants modify only the parameters noted in the text below.}
\label{tab:base-hyperparams}
\begin{tabular}{l c}
\toprule
\textbf{Parameter} & \textbf{Value} \\
\midrule
Learning rate & 0.001 \\
Buffer size & 1000000 \\
Learning starts & 5000 \\
Batch size & 128 \\
Target smoothing coefficient $\tau$ & 1.0 \\
Discount factor $\gamma$ & 0.99 \\
Train frequency & 16 \\
Gradient steps & 4 \\
Target update interval & 2000 \\
Exploration fraction & 0.1 \\
Initial $\epsilon$ & 1.0 \\
Final $\epsilon$ & 0.1 \\
Max grad norm & 10 \\
Intrinsic reward scaling $\beta$ & 0.1 (Multi-NGU variants only) \\
\bottomrule
\end{tabular}
\end{table}

For the Multi-DQN baseline, the configuration is identical to Table~\ref{tab:base-hyperparams} except that the intrinsic reward scaling $\beta$ does not apply. 
In the shared novelty variant, we additionally introduced the novelty threshold parameter $k$, with $k=1$. 
In the heterogeneous $\beta$ variant, different agents were assigned distinct values of $\beta$ with $\{0.1,0.2,0.4\}$.

To determine the final configuration, we conducted a grid search over key hyperparameters of the Multi-DQN baseline, varying batch size $\{64,128\}$, training frequency $\{1,4,16\}$, gradient steps $\{1,4,8\}$, and target update interval $\{100,500,1000\}$, while keeping other values fixed. The best-performing setup was consistent with the configuration reported in Table~\ref{tab:base-hyperparams}, with two exceptions: the target update interval was set to $2000$ steps, and the $\epsilon$-greedy schedule was annealed from $1.0$ to $0.1$. 

For Multi-NGU, we additionally tuned the intrinsic reward scaling parameter $\beta \in \{0,0.1,0.5,1.0\}$, finding that $\beta=0.1$ yielded the most stable learning and highest average returns.

\section{Multi-novelty Analysis}
\label{sec:appendix_multinovelty}

\begin{figure}[htpb]
    \centering
    \includegraphics[width=0.95\linewidth]{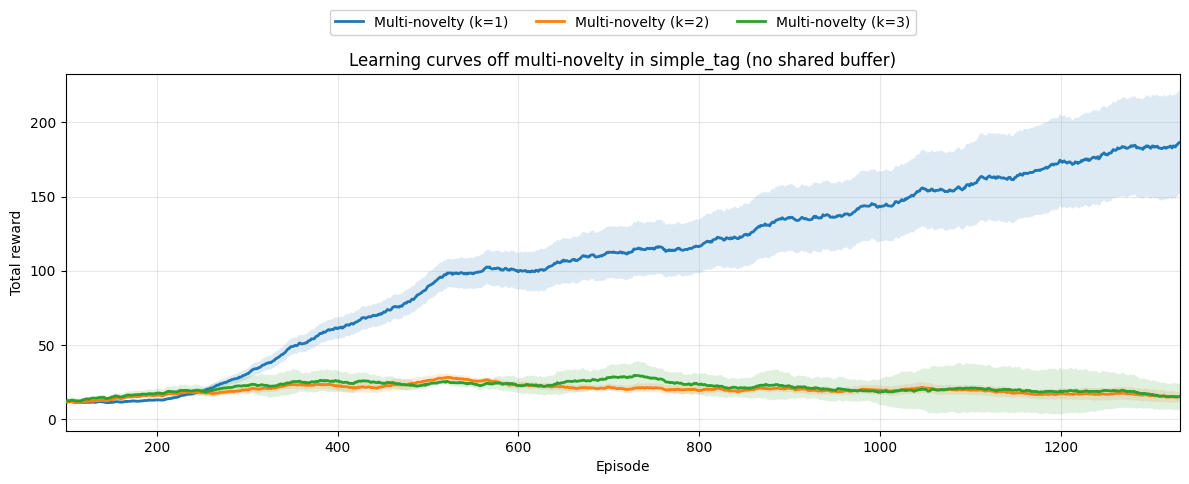}
    \caption{Results for $k=1$ and $k=2$ are averaged over 15 runs, while $k=3$ is averaged over 5 runs. All curves are smoothed with a window of 100, and the shaded regions indicate the 95\% confidence interval.}
    \label{fig:appendix_multinovelty}
\end{figure}

To further investigate the effect of the novelty-sharing threshold, we evaluated the multi-novelty variant with $k=1,2,3$. Results are shown in Figure~\ref{fig:appendix_multinovelty}. Consistent with the main text, $k=1$ provides the best performance, achieving stable learning and substantially higher returns compared to $k=2$ and $k=3$. Increasing $k$ beyond $1$ leads to degraded performance, as novelty signals become less informative when averaged over multiple neighbors. We report that the curve for $k=3$ is based on only 5 runs due to computational constraints, whereas the other settings were run 15 times. Nevertheless, the trend is clear: higher $k$ values reduce the effectiveness of intrinsic rewards for exploration.

\section{Heterogeneous $\beta$ Analysis}
\label{sec:appendix_hetero_beta}

In the heterogeneous $\beta$ variant, we evaluated multiple assignments across agents, such as $(\beta_1=0.1,\;\beta_2=0.2,\;\beta_3=0.4)$, $(\beta_1=0.0,\;\beta_2=0.3,\;\beta_3=1.0)$, and $(\beta_1=0.0,\;\beta_2=0.1,\;\beta_3=0.5)$. Among these, only the first configuration provided consistent improvements over the baseline.

\begin{figure}[htpb]
    \centering
    \includegraphics[width=0.95\linewidth]{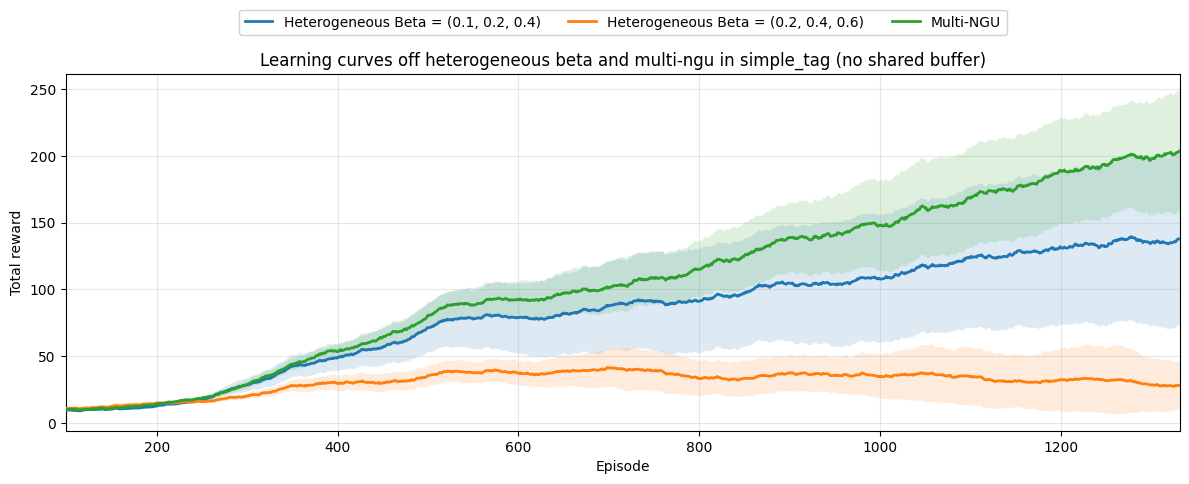}
    \caption{Learning curves of heterogeneous $\beta$ variants compared with Multi-NGU in the simple\_tag environment. Results are averaged over 15 runs with smoothed returns (window=100), except for $(0.2, 0.4, 0.6)$ which is averaged over 10 runs. The shaded regions indicate the 95\% confidence interval.}
    \label{fig:appendix_hetero_beta}
\end{figure}

Figure~\ref{fig:appendix_hetero_beta} shows the results of the heterogeneous $\beta$ variants compared against Multi-NGU. We report two heterogeneous configurations: $(0.1, 0.2, 0.4)$ and $(0.2, 0.4, 0.6)$. While both underperform compared to standard Multi-NGU, the configuration with smaller $\beta$ values exhibits more stable learning and higher returns than the larger set. We report that the configuration $(0.2, 0.4, 0.6)$ was run for only 10 seeds due to computational constraints, while the other setups used 15 runs. Overall, these experiments suggest that heterogeneous $\beta$ values do not yield consistent improvements over a small common $\beta$.

\end{document}